\documentclass[runningheads]{llncs}
\usepackage{comment}
\usepackage{booktabs}
\usepackage{algorithm} 
\usepackage{algpseudocode} 
\usepackage{caption}
\usepackage{graphicx}
\usepackage{multirow}
\usepackage{cite}
\usepackage{mathtools,amssymb}
\usepackage{subcaption}
\usepackage{pgfplots}
\usepackage{hyperref}
\usepackage{xcolor}
\usepackage{soul}
\usepackage{svg}

 \usepackage{multirow}
\begin{document}
%
\title{DiffDefense: Defending against Adversarial Attacks via Diffusion Models \thanks{This work was supported by the European Commission under European Horizon 2020 Programme, grant number 951911—AI4Media}}
%
%
%
\author{Hondamunige Prasanna Silva \and Lorenzo Seidenari \and Alberto Del Bimbo }

\authorrunning{Silva H. P., Seidenari L., Del Bimbo A.}
%
\institute{hondamunige.silva@gmail.com, lorenzo.seidenari@unifi.it, alberto.delbimbo@unifi.it \\ University of Florence, Italy}
%
\maketitle              
\begin{abstract}
This paper presents a novel reconstruction method that leverages Diffusion Models to protect machine learning classifiers against adversarial attacks, all without requiring any modifications to the classifiers themselves. The susceptibility of machine learning models to minor input perturbations renders them vulnerable to adversarial attacks. While diffusion-based methods are typically disregarded for adversarial defense due to their slow reverse process, this paper demonstrates that our proposed method offers robustness against adversarial threats while preserving clean accuracy, speed, and plug-and-play compatibility. 

Code at: \href{https://github.com/HondamunigePrasannaSilva/DiffDefence}{https://github.com/HondamunigePrasannaSilva/DiffDefence}.

\keywords{Diffusion models  \and Adversarial defense \and Adversarial Attack.}
\end{abstract}

\section{Introduction}

\begin{figure}[htb]
\centering
\includegraphics[width=.95\textwidth]{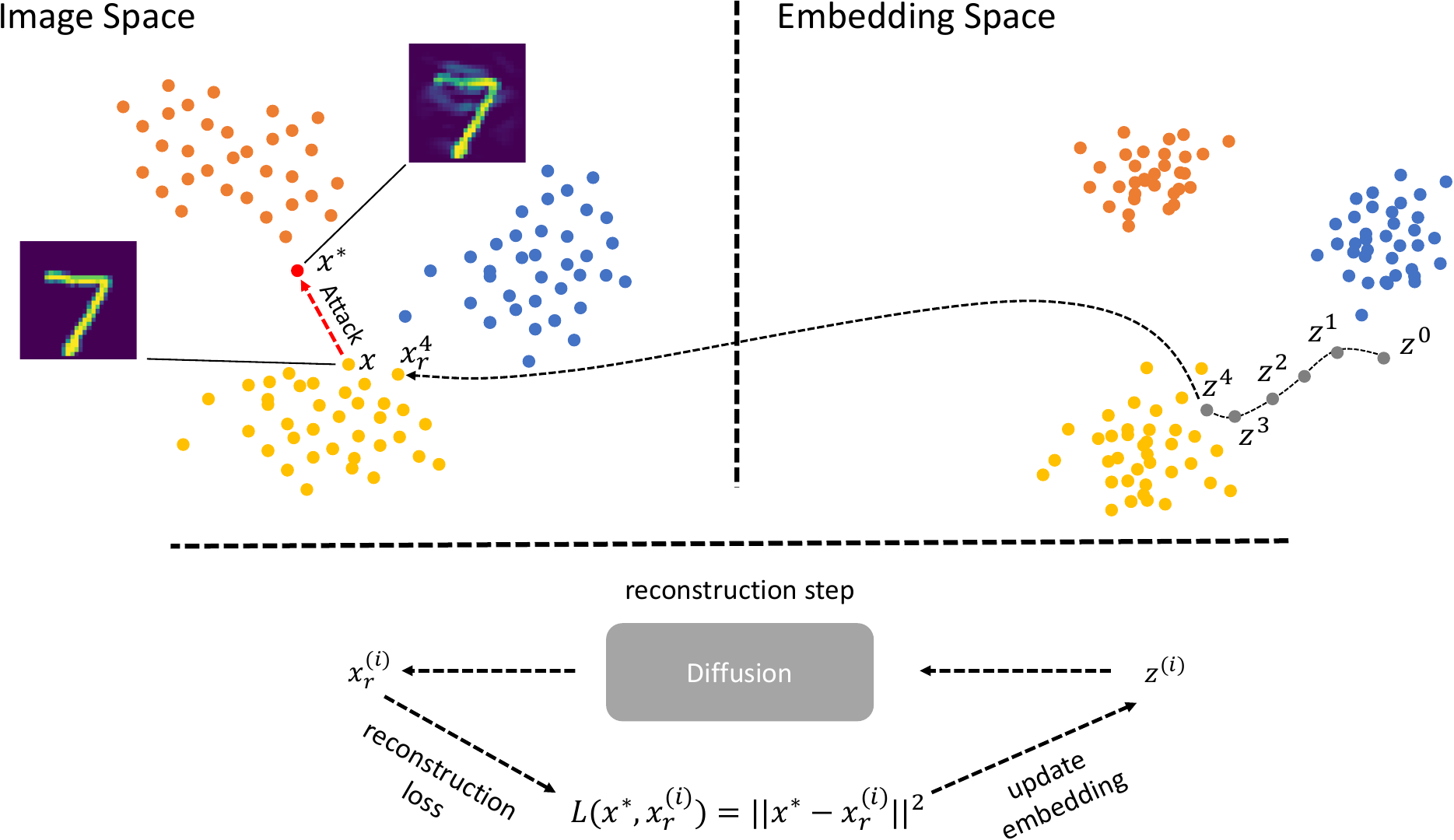}

\caption{Overview of our approach. Adversarial attacks happen in image space by adding crafted noise to a pattern $x$, shifting classifier's output to a wrong class. DiffDefence starts by drawing a sample $z^{1}_T$, to diffuse iteratively, for $T$ steps into a reconstruction $x^{(i)}_r = z^{(i)}_0 = G\left(z^{(i)}_T\right)$; we then optimize $z^{(i)}_T$ so that the diffusion output for a given optimized pattern $z_T^{(i+1)}$, lies closer to the original attacked sample. In the figure, we drop the diffusion step subscript for readability purposes \label{fig:arch}.}
\end{figure}

The susceptibility of machine learning models to adversarial attacks is a major challenge in the field of artificial intelligence. While various techniques have been proposed to enhance the robustness of classifiers against such attacks, there is a pressing need for more effective and efficient solutions. In recent years, generative models such as Generative Adversarial Networks (GANs)\cite{gan}, Diffusion Probabilistic Models\cite{diffusionDDPM} have emerged as a promising approach to improve the resilience of machine learning models against adversarial attacks.

\begin{figure}[!htb]
\centering
\includegraphics[width=.70\textwidth]{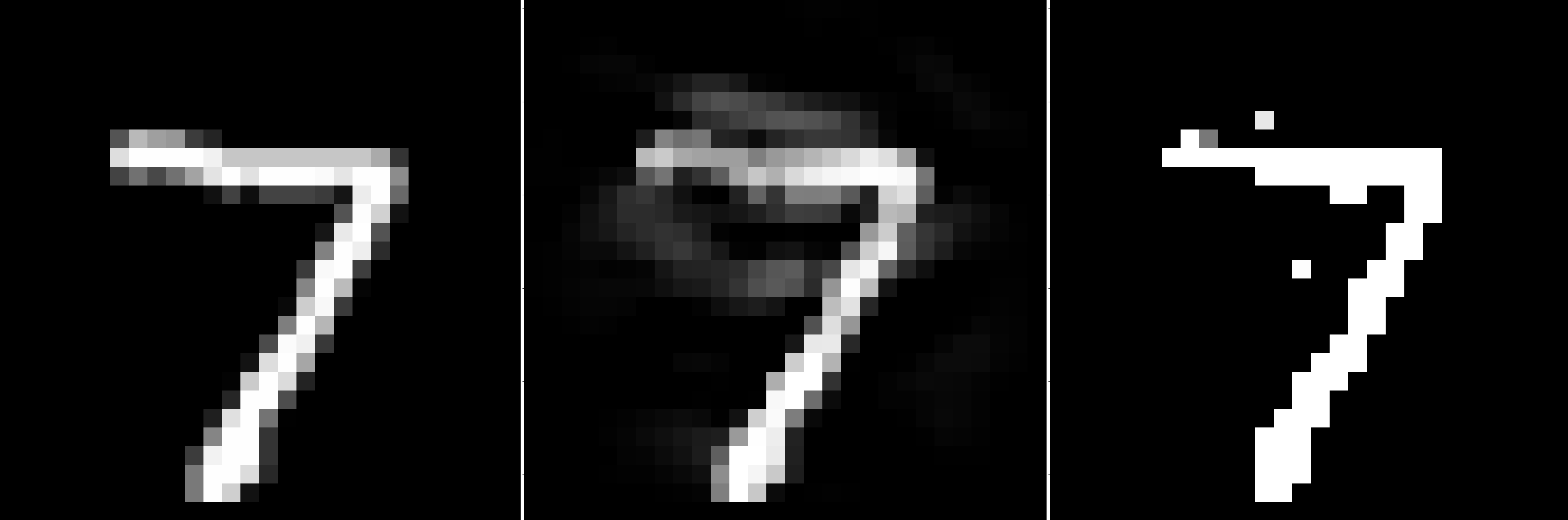}
\caption{Left original example from MNIST. Center result of the DeepFool attack (wrongly classified). Right our DiffDefense reconstruction (correctly classified).      \label{fig:imgs}
}
\end{figure}

Modern deep generative models have a common structural similarity: the generation of novel patterns is usually performed by transforming some random latent code $z$. Sampling $z \sim p(z)$, where $p(z)$ is a known distribution (e.g. $\mathcal{N}(0,I)$ and then computing $G(z)$, where $G(\cdot)$ is a deep neural network allows the generation of new data. Given a model $G(\cdot)$, trained on clean data we can assume that attacked samples $x^*$ have a different distribution, therefore finding a latent code $z^*$ able to generate $x^*$ should be hard. Our approach builds on the idea that given some attacked pattern $x^* = x + \epsilon$ where $x$ is an unknown clean pattern and $\epsilon$ is a perturbation crafted to induce some classifier into a mistake, we should be able to find some latent code $z^*$ for which $G(z^*)$ is closer to the unknown clean pattern $x$ than to the attacked one $x^*$.  In Fig.~\ref{fig:imgs} it can be seen how an attack can add subtle patterns (center) to a clean image(left) and how DiffDefense recover a correctly classified example(right).

In this paper, we present a novel approach that leverages Diffusion Models to enhance the resistance of machine learning classifiers to adversarial attacks. Our proposed method involves reconstructing the input image using a reverse process of a diffusion model (see Fig.~\ref{fig:arch} for details), which improves the model's ability to withstand adversarial attacks. We show that this approach offers comparable speed and robustness to other generative model-based solutions. Moreover, our proposed defense mechanism can be applied as a plug-and-play tool to any classifier without compromising its accuracy, provided that the diffusion model can generate high-quality images.
Overall, our approach holds promise as a viable alternative to other more complex to train models, such as GANs, for defending against adversarial attacks on machine learning models, owing to the benefits offered by Diffusion Models. 

Our contribution is threefold:
\begin{itemize}
\item We are the first to use recently successful Denoising Diffusion Probabilistic models  as a plug-in algorithm for reconstruction based adversarial defense. Differently from \cite{diffguidenoise, diffpure} our approach is based on reconstruction thus not requiring backward and forward passes for each optimization step. Moreover, DDPMs are more stable in training with respect to GANs which have also been used as a reconstruction tool\cite{DefenceGan}.

\item Thanks to a superior reconstructive and representational power, DDPMs require less prototype embeddings and iterations to extract a clean pattern from the attacked one, leading to higher inference efficiency with respect to \cite{diffguidenoise, diffpure, DefenceGan}.

\item Finally, our approach does not require to be trained on adversarial patterns and can be used to detect attacked images.
\end{itemize}

\vspace{-30pt}
\section{Related Works}

We now cover existing state of the art on modern generative models in adversarial machine learning, forming the base of our reconstruction based adversarial defense approach. We than discuss recent methods of adversarial attacks and defense.

\subsection{Generative Models}
Generative models\cite{kingma2019introduction, gan, diffusionDDPM, ddim} have emerged as a powerful class of machine learning algorithms that can create new data samples with characteristics similar to a given dataset. Their central idea is to learn the underlying probability data distribution, which could then be used to generate new patterns via sampling.  Interestingly, these models have also proven to be particularly effective in adversarial scenarios, showcasing their ability to create samples that can attack classifiers\cite{advgan, atgan}. Aforementioned models have also shown the capability to learn a semantically coherent latent embedding space. This property has been exploited, in adversarial scenarios, to remove attacks from patterns. By reconstructing\cite{DefenceGan, defenceVae}, purifying \cite{diffpure} the perturbed sample and generating new samples to bolster adversarial training\cite{gowal21}, these models can significantly enhance the security of machine learning systems. Recent research has highlighted the potential role of generative models in adversarial defense, as their primary objective is to produce fake data that closely resembles real data. 
In this paper we want to investigate the use of Diffusion Models to bolster the robustness of models against adversarial attacks. For a  more thorough coverage of modern Deep Generative Models we refer the reader to~\cite{bond2021deep}.

\subsection{Adversarial Attacks and Defense}


An adversarial attack is a process that aims at altering a classifier input pattern in order to get the classifier to output a wrong prediction. So given an input $x_i$ and a corresponding label $y_i$, and a classifier $\mathcal{C}$ an attack method will aim at obtaining a $x^*$ such that $\mathcal{C}(x^*) \neq y_i$. Given this definition any method completely replacing pattern $x$ with a different pattern would suffice in making a classifier mistake the label. For this reason, a constraint on the perturbation of the attacked pattern $x^*$ is also required. Therefore the attacked pattern $x^*$ must be close to the original one $||x-x^*|| < \epsilon$.

Adversarial attacks can work in white-box and black-box scenarios. In the white-box scenario the classifier is known to the attacker. The full knowledge of the classification method implies that, for example for a neural network all weights are known and the attacking method can leverage this knowledge. White-box approaches may exploit the computation of the gradient of the model loss with respect to its parameters for a specific input, such as the Fast Gradient Signed Method (FGSM)\cite{FGSM}. PGD \cite{pgd, advtrain} improves over \cite{FGSM} by a refined attack generation obtained by iterative Projected Gradient Descent (PGD). Instead, Deepfool\cite{deepfool} attempts to find the closest decision boundary to then perturb the input in that direction. Combination of attacks have been also proposed in~\cite{autoattack}, combining parameter free versions of PGD with SquareAttack\cite{squarea}. EOT+PGD combines the concepts of Expectation over Transformation (EOT)\cite{athalye2018synthesizing} and Projected Gradient Descent (PGD) to improve its effectiveness. Elasticnet~\cite{elasticnet} exploits a combination of L1 and L2 regularisation terms to provide an optimal trade-off.

In the black-box scenario, the attacking method does not have access to the classifier, which is the most realistic setting. When an attacker has no knowledge of a classifier's architecture and weights, they can employ a query-based approach to perturb the input without relying on the gradient by applying a perturbation on the input until the classifier changes its output. Depending on the model’s feedback for a given query, an attack can be classified as a score-based\cite{chen2017zoo} or hard-label\cite{brendel2017decision} attack. In the score-based setting,the attacker exploits the model's output probabilities of each decision. Several attacks have been created with this approach, such as Square Attack\cite{squarea}, which selects localized square-shaped updates at random positions. Pixel Attack \cite{su2019one} show attacks are possible even with a single pixel perturbation. 

In the hard label-based approach, the attacker exploits the model's final decision output. Recently, SIGN-OPT\cite{cheng2019sign} improved a previous work \cite{cheng2018query} using fewer queries (20K) to attack, being faster than the previous and obtaining a similar error rate of white-box attacks, but remaining much slower than the white-box.
However, a query-based approach may not be as efficient as a white-box attack. Instead, a black-box attack can be employed by using the transferability\cite{papernot2016transferability, papernot2017practical} of perturbed images to attack the target model. The attacker can use a substitute model, where they have full knowledge, to generate adversarial images using white-box attacks that can then be used to attack the target model. Further coverage of adversarial attack and defense technique can be found in \cite{akhtar2018threat}.

The issue of adversarial attacks in machine learning has prompted the development of various defense mechanisms, we can roughly identify three main approaches: adversarial training, adversarial reconstruction, and adversarial purification. Adversarial training methods improve the robustness of a model against adversarial attacks by augmenting its training set with adversarial examples. Introduced first by \cite{FGSM}, adversarial training has become one of the most successful defense against adversarial attacks \cite{advtrain, gowal20, rebuffi21}, adversarial training can also be enhanced using generative models for data augmentation \cite{gowal21}. The main limitation of this approach is that is mainly protecting classifiers from methods used in the adversarial training.

Adversarial reconstruction approaches leverage the projection of patterns onto a learned latent manifold to regenerate the original input from its adversarial counterpart.  Generative models~\cite{DefenceGan, defenceVae} are a natural choice to learn such latent representation and to obtain clean reconstructions out of attacked patterns. Other approaches leveraged super-resolution networks\cite{mustafa2019image} or trained a reconstruction network to minimize the perceptual loss between the reconstruction of the attacked pattern and the clean image. 

Adversarial purification techniques perform a filtering of attacked patterns removing adversarial perturbations while preserving their original features. Recently proposed denoising Diffusion Models have been used as a tool for purification~\cite{advpursb, diffguide, diffguidenoise}

Reconstructions based on GANs\cite{DefenceGan} are effective and generalize versus unseen threats. However, the instability of GANs during training remains a challenge. Moreover, many source noise embeddings and multiple reconstruction iterations are required to obtain effective defense. Defense-VAE is faster and as effective as Defense-GAN. However, to obtain effective reconstructions \cite{defenceVae} the method is fine-tuned on attacked images making the approach less general and more prone to fail on unseen threats. Purification via Diffusion Models\cite{diffpure, diffguide, diffguidenoise} exploits multiple forward/backward passes to obtain a reliable defense which requires significant time to purify an image.

With respect to \cite{defenceVae, DefenceGan} our approach exploits powerful Diffusion Models as a reconstruction tool. Interestingly, our approach is more efficient then Defense-GAN, requiring less iterations and source embeddings. Different from\cite{defenceVae} we do not require to train on adversarial examples to work as Defense-VAE\cite{defenceVae}. Current defense mechanisms exploiting Diffusion models are less efficient, requiring as much as 5s on a V100 card\cite{diffpure}, while our approach runs in 0.28s on a TitanXP card.



\section{Methodology}

We propose a diffusion reconstruction method as a defense against adversarial attacks. The underlying idea is that adversarial attacks seek to deceive a deep neural network (DNN) by introducing a disturbance to the image while preserving its semantic meaning. Hence, the adversarial image ought to be situated close to the original, unperturbed image.

Our approach is based on the idea that it is possible to induce a Generative model $G(\cdot)$ to produce a given image $x^*$ by minimizing the distance in image space of the output pattern, getting $\hat{z}$ as the result of such minimization 
\begin{equation}
    \hat{z} = \arg \min_{z} ||G(z) - x^*||
    \label{eq:reco}
\end{equation}
Obtaining the reconstructed image $x_r = G(\hat{z}$). 
Having $G(\cdot)$ being learned on a clean dataset, the main assumption is that output generated obtained by solving Eq.~\ref{eq:reco} are closer to clean examples than corrupted ones. 

In our case $G(z)$ is the result of a reverse diffusion process, each step of which is given by: \begin{equation}
\centering
z_{t-1} = \frac{1}{\sqrt{\alpha_{t}}}- \left(z_{t} - \frac{1-\alpha_{t}}{\sqrt{1- \bar{\alpha_{t}}}} \epsilon_{\theta}\left(z_{t}, t\right)\right) + \sigma_{t}n
    \label{eq:reverse}
\end{equation}

where $\epsilon_{\theta}$ is the U-Net noise prediction model,  $\alpha_{t} = 1- \beta_{t}, \bar{\alpha_{t}} = \prod_{s=1}^{t} \alpha_{s},\{\beta_{t} \in (0,1)\}_{t=1}^{T}, \sigma_{t} = \sqrt{\beta_{t}}$.



Our goal is to have the diffusion reverse process create a clean image, that is as close as possible to the attacked input. To this end we must obtain a suitable noise vector $z_k$. Therefore, we start from a random noise sample $z_T^1$, and we iteratively generate an image using the reverse process of a diffusion model. We then optimize z to solve Eq.~\ref{eq:reco} as shown in Algorithm~\ref{alg:reconstruct}. In its general form, the proposed algorithm may also be run for multiple source embeddings, although we found that it only increase slightly the accuracy. 



\begin{algorithm}[h]
	\caption{DiffDefense Reconstruction Algorithm. As a loss $\mathcal{L}(x_r^{(i)},x^*)$ we used Mean Square Error (MSE). $T^*$ are the diffusion steps and $L$ are the gradient descent iterations, both treated as hyperparameters. $\Delta=0.1$ is a decay rate. } 
        \label{alg:reconstruct}
	\begin{algorithmic}[1]
            
    	\State Given adversarial image $x^*$
            \State $z^1_T \sim \mathcal{N}(0,\,I)$
		\For {$i=1,2,\ldots,L$}
                
			\For {$t=T^*,T^*-1,\ldots,0 \ steps$}
                    \State $n \sim \mathcal{N}(0,\,I)$
				\State  $z^{(i)}_{t-1} = \frac{1}{\sqrt{\alpha_{t}}}- \left(z^{(i)}_{t} - \frac{1-\alpha_{t}}{\sqrt{1- \bar{\alpha_{t}}}} \epsilon_{\theta}\left(z^{(i)}_{t}, t\right)\right) + \sigma_{t}n $
			\EndFor
                
                \State $\eta^{i} = \eta^{i-1} \Delta^{\frac{1}{\left \lceil L * 0.8 \right \rceil}}$
                \State $x_r^{(i)} = z_0^{(i)}$
			\State $z_T^{(i + 1)} = z_T^{(i)}  - \eta^{i} \nabla_z \mathcal{L}(x_r^{(i)}, x^*)$
		\EndFor
	\end{algorithmic}

\end{algorithm}
\vspace{-20pt}
\subsection{Implementation details}
The noise prediction U-Net $\epsilon_{\theta}$ architecture consists of a contracting path, bottleneck layer and an expansive path. The contracting path involves repeating a block with layer normalization, 3x3 convolutions, and SiLU activation, followed by downsampling with a stride of 2. The number of feature channels is doubled at each of the three downsampling step. The bottleneck layer consists of the same block of the contracting path repeated three times. The expansive path starts by concatenating the corresponding feature map from the contracting path with an upsampled input using transpose convolution. It is then followed by a block with layer normalization, 3x3 convolutions, and SiLU activation. At each upsampling step, the number of feature channels is halved. This process is also repeated three times. Both the contracting and expansive paths include a time-embedding layer, which consists of two linear layers with a SiLU activation in between. This time-embedding layer is added at each block of the contracting and expansive paths.

We employ two classifiers: the attacked classifier A and the surrogate classifier B. Classifier A is composed of two 5x5 convolutions with 64 output channels and stride of 2 and 1, respectively, using ReLU activations. It is then followed by a dropout layer (p = 0.25), a linear layer with 128 output features using ReLU activation, another dropout layer (p = 0.5), and finally a linear layer with 10 output features.
We use classifier B to generate adversarial samples for black box attacks. B consists of a dropout layer (p = 0.2), followed by three convolutions with respective filter sizes of 8x8, 6x6, and 5x5 and strides of 2, 2, and 1, using ReLU activations. Afterward, another dropout layer (p = 0.5) is applied, and the final layer is a linear layer with 10 output features.

The Diffusion Model and the classifier are trained using the same clean dataset. Training the classifier on reconstructed data is unnecessary if the diffusion model generates high-fidelity images resembling the originals. 

To implement the adversarial attack used to evaluate DiffDefense, we used adversarial robustness toolkit\cite{art} and torchattacks\cite{torchattacks}.

\section{Experiments}
This section presents the experiments that evaluate the proposed method using both black-box and white-box attacks. First, we evaluate performance against three classic attacks in both settings\cite{FGSM, pgd, deepfool}. In these experiments we seek optimal values for the number of iterations for gradient descent $L$, the embedding set size $R$, and the diffusion step $T^*$. Then, keeping these hyperparameter fixed we test DiffDefense against unseen attacks~\cite{squarea, autoattack, eotpgd, elasticnet}. Finally, we use this method to detect adversarial samples. The experiments are conducted on two different datasets, MNIST\cite{mnist} and KMNIST\cite{kmnist}.
\begin{table}[!htb]
  \caption{ Performance of DiffDefense on white box \& black box attacks on MNIST \& KMNIST datset. We report accuracy for each attack with and without defense. For Black blox attacks, adversarial images has been crafted using a substitute classifier. }
  \centering
  \addtolength\tabcolsep{10pt}
  \renewcommand{\arraystretch}{1.2}
  \resizebox{.9\textwidth}{!}{
  \begin{tabular}{c|cccc}
    \toprule
    Dataset  & Attack & Type & Without defense    & With defense  \\
    \midrule
        \multirow{7}{4em}{MNIST} 
                                    & No attack & - & $99.14\%$ & $99.06\%$ \\ \cline{2-5}
                                    & \multirow{2}{4em}{DeepFool}
                                                                  & White box & $0.95\%$  & $98.16\%$ \\
                                                                  &  & Black box & $97.17\%$  & $98.86\%$ \\\cline{2-5} 
                                    & \multirow{2}{4em}{PGD}
                                                                  & White box & $5.81\%$  & $95.94\%$ \\
                                                                  &  & Black box & $51.28\%$  & $97.18\%$ \\ \cline{2-5}
                                    & \multirow{2}{4em}{FGSM}
                                                                  & White box & $23.72\%$ & $89,95\%$ \\
                                                                  &  & Black box & $15.81\%$ & $91.28\%$ \\ 
        \hline
         \multirow{7}{4em}{KMNIST} 
                                    & No attack & - & $95.18\%$ & $94.38\%$ \\ \cline{2-5}
                                    & \multirow{2}{4em}{DeepFool}
                                                                  & White box  & $2.93\%$  & $93.92\%$ \\
                                                                  &  & Black box & $92.16\%$  & $93.92\%$ \\\cline{2-5} 
                                    & \multirow{2}{4em}{PGD}
                                                                  & White box & $26.83\%$  & $84.85\%$ \\
                                                                  &  & Black box & $58.43\%$  & $91.49\%$  \\ \cline{2-5}
                                    & \multirow{2}{4em}{FGSM}
                                                                  & White box& $37\%$ & $79.5\%$ \\
                                                                  &  & Black box& $49.96\%$ & $88.53\%$ \\   
    \bottomrule
  \end{tabular}}
  \label{fig:wh_bb_tab}
\end{table}
\subsection{Result of white-box \& black-box attack} We  investigate DiffDefense ability to withstand both white-box and black-box attacks. To this end, we subject it to three potent white-box attacks: FGSM, PGD, and Deep Fool. Furthermore, we evaluate the performance of DiffDefense against these same attacks in the black-box setting, where we generate adversarial samples using an auxiliary classifier to attack the target classifier. In Tab. \ref{fig:wh_bb_tab} we can see how all attacks are pretty effective in both settings except for DeepFool used as a black box method. In general using DiffDefense we can always recover a correct classification for almost all attacked examples.

\begin{table}[!htb]
\vspace{-20pt}
  \caption{ Robustness of DiffDefense against unseen threats on MNIST dataset. Adversarial training using adversarial sample crafted by FGSM attack $\epsilon=0.3$.  }
  \label{result}
  \centering
  \addtolength\tabcolsep{4.5pt}
  \resizebox{.9\textwidth}{!}{
  \begin{tabular}{lcccc}
    \toprule
    Attack & Type & W/O Defense    & W/ Defense & Adv. Training \\
    \midrule
        FGSM       $\epsilon=0.3$    & White Box & $23.72\%$ & $89.95\%$   & $98.02\%$   \\
        PGD       $\epsilon=0.3$     & White Box & $5.81\%$  & $95.94\%$   & $79.59\%$   \\ 
        Deep Fool                    & White Box & $0.95\%$  & $98.16\%$   & $5.81\%$    \\
        EOT+PGD       $\epsilon=0.3$ & White Box & $24.57\%$ & $96.46\%$   & $89.22\%$   \\
        Square Attack                & Black Box & $43.31\%$ & $97.31\%$   & $93.09\%$   \\
        AutoAttack                   & White Box & $1.26\%$  & $88.09\%$   & $45.86\%$   \\
        Elastic Net                  & White Box & $0.75\%$  & $95.75\%$   & $0.62\%$    \\
    \bottomrule
  \end{tabular}}
  \label{fig:newattacks_tab}
\end{table}
\subsection{Defense against unseen threats}
One of the significant limitations of existing adversarial training defense methods is their inability to effectively address previously unseen threats. DiffDefense does not require to observe adversarial patterns to work, nonetheless previous experiments were performed seeking the optimal values for hyperparameters $L, R, T^*$.  In order to assess the robustness of our proposed approach to such unseen attacks we conducted evaluations using four different attack techniques: Square Attack\cite{squarea}, Auto Attack\cite{autoattack} , EOT+PGD\cite{eotpgd} attack and Elastic Net \cite{elasticnet}, without tuning hyperparameters. Results in Tab.~\ref{fig:newattacks_tab}  indicate that our method is robust against all four of these previously unseen attack methods.  Here we also test the behavior of adversarial training with samples produced by FGSM with $\epsilon=0.3$. Interestingly, DiffDefense obtains high accuracy even in cases in which  adversarial training, is not helping at all\cite{deepfool, autoattack, elasticnet}.
\subsection{Ablation Studies}
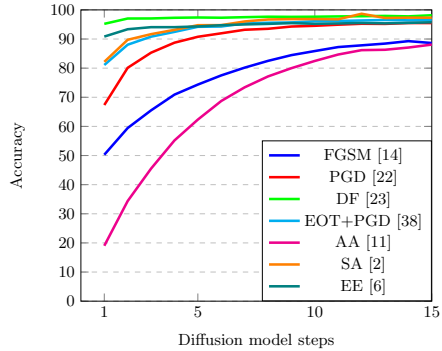
\begin{figure}[!htb]
    \centering
    \resizebox{.5\textwidth}{!}{
    \begin{tikzpicture}[scale = 0.6]
    \begin{axis}[
        xlabel={Diffusion model steps}, ylabel={Accuracy},
        xmin=0, xmax=15,ymin=0, ymax=100,
        xtick={1,5,10,15},
        ytick={0,10,20,30,40,50,60,70,80,90,100},
        legend style={at={(1,0)},anchor=south east},
        ymajorgrids=true,
        grid style=dashed,
    ]
    
    \addplot[color=blue, line width=0.5mm]
        coordinates {(1,50.27)(2,59.44)(3,65.51)(4,70.91)(5,74.36)(6,77.51)(7,80.17)(8,82.5)(9,84.47)(10,85.87)(11,87.24)(12,87.83)(13,88.43)(14,89.27)(15,88.63)};
    \addlegendentry{FGSM \cite{FGSM}}
    
    \addplot[color=red, line width=0.5mm]
        coordinates {(1,67.34)(2,80.12)(3,85.31)(4,88.74)(5,90.77)(6,91.96)(7,93.18)(8,93.51)(9,94.29)(10,94.54)(11,94.92)(12,95.24)(13,95.19)(14,95.49)(15,95.48)};
    \addlegendentry{PGD\cite{pgd}}
    
    \addplot[color=green, line width=0.5mm]
        coordinates {(1,95.21)(2,97.04)(3,97.06)(4,97.29)(5,97.39)(6,97.31)(7,97.5)(8,97.66)(9,97.61)(10,97.66)(11,97.76)(12,97.83)(13,97.93)(14,97.78)(15,98.16)};
    \addlegendentry{DF\cite{deepfool}}

    \addplot[color=cyan, line width=0.5mm]
        coordinates {(1,81.08)(2,88.05)(3,90.81)(4,92.45)(5,94.19)(6,94.29)(7,95.39)(8,95.58)(9,95.7)(10,96.11)(11,96.18)(12,96.37)(13,96.54)(14,96.41)(15,96.46)};
    \addlegendentry{EOT+PGD\cite{eotpgd}}

    \addplot[color=magenta, line width=0.5mm]
        coordinates {(1,19.02)(2,34.38)(3,45.51)(4,55.18)(5,62.34)(6,68.72)(7,73.41)(8,77.16)(9,79.99)(10,82.47)(11,84.68)(12,86.18)(13,86.31)(14,87.07)(15,88.09)};
    \addlegendentry{AA\cite{autoattack}}
    
    \addplot[color=orange, line width=0.5mm]
        coordinates {(1,82.16)(2,89.74)(3,91.67)(4,93.15)(5,94.75)(6,94.77)(7,96.03)(8,96.71)(9,96.83)(10,96.85)(11,96.78)(12,98.68)(13,97.22)(14,97.33)(15,97.31)};
    \addlegendentry{SA\cite{squarea}}
    
    \addplot[color=teal, line width=0.5mm]
        coordinates {(1,90.84)(2,93.36)(3,94.07)(4,94.07)(5,94.32)(6,94.74)(7,95.01)(8,95.2)(9,95.46)(10,95.22)(11,95.35)(12,95.51)(13,95.38)(14,95.54)(15,95.75)};
    \addlegendentry{EE\cite{elasticnet}}
        
    \end{axis}
    
    \end{tikzpicture}  }
    \caption{Accuracy analysis of the classifier after DiffDefense has been applied on different diffusion steps. Using L = 4 and R = 5.}
    \label{fig:diffusion_steps}
\vspace{-20pt}
  
\end{figure}

To evaluate the effectiveness and speed of our proposed approach, we conducted an analysis of the three main hyperparameters, iteration number $L$, embedding set size $R$ and  diffusion step $T^*$.
We found  that the proposed method does not need the same amount of steps of the Diffusion Model but it converges with less steps, as shown in Fig.\ref{fig:diffusion_steps}. Moreover, in a comparison with Defense-GAN\cite{DefenceGan}, the results of our experiments revealed that our method achieved convergence with fewer iteration steps and a smaller embedding set, while also requiring less time to converge than the GAN-based method. This was evident in the results presented in Tab.\ref{fig:mytable1}, which show the superiority of our proposed approach over Defense-GAN. The metrics used for comparison include robust accuracy, which measures the accuracy after applying the defense, and time, which indicates the duration to reconstruct a single image.
\vspace{-20pt}
\begin{table}[H]
\centering
      \caption{Comparison with Defense-GAN\cite{DefenceGan}. Fewer iterations (L) and smaller embedding set (R) in DiffDefense lead to faster convergence and reduced time. All tests made on MNIST using white-box FGSM attack ($\epsilon=0.3$) and the same classifier as \cite{DefenceGan}}
       \addtolength\tabcolsep{19pt}
       \resizebox{.9\textwidth}{!}{
       \begin{tabular}{c|rrcc}
        \toprule
            Method  & L & R & Time & Robust Acc  \\
            \midrule
                \multirow{4}{5em}{Defense Gan\cite{DefenceGan}} 
                & 25    & 10 & 0.086  & $79.98\%$ \\
                & 100    & 1  & 0.273 &$50.11\%$ \\
                & 100    & 10 & 0.338  &$89.11\%$ \\
                & 200    & 10 & 0.675  & $91.55\%$ \\
                \hline
                \multirow{2}{4em}{Ours} 
                & 5    & 1 & 0.280  & $87.78\%$ \\
                & \textbf{5}   & \textbf{5} & \textbf{0.280}  & $\textbf{89.95}\%$ \\ 
                        
            \bottomrule
        \end{tabular}}
        \label{fig:mytable1}
        \vspace{-15pt}
\end{table}

\subsection{Attack detection}
\begin{figure}[H]
\vspace{-25pt}
     \centering
     \begin{subfigure}[b]{0.43\textwidth}
         \centering
         \includegraphics[width=\textwidth]{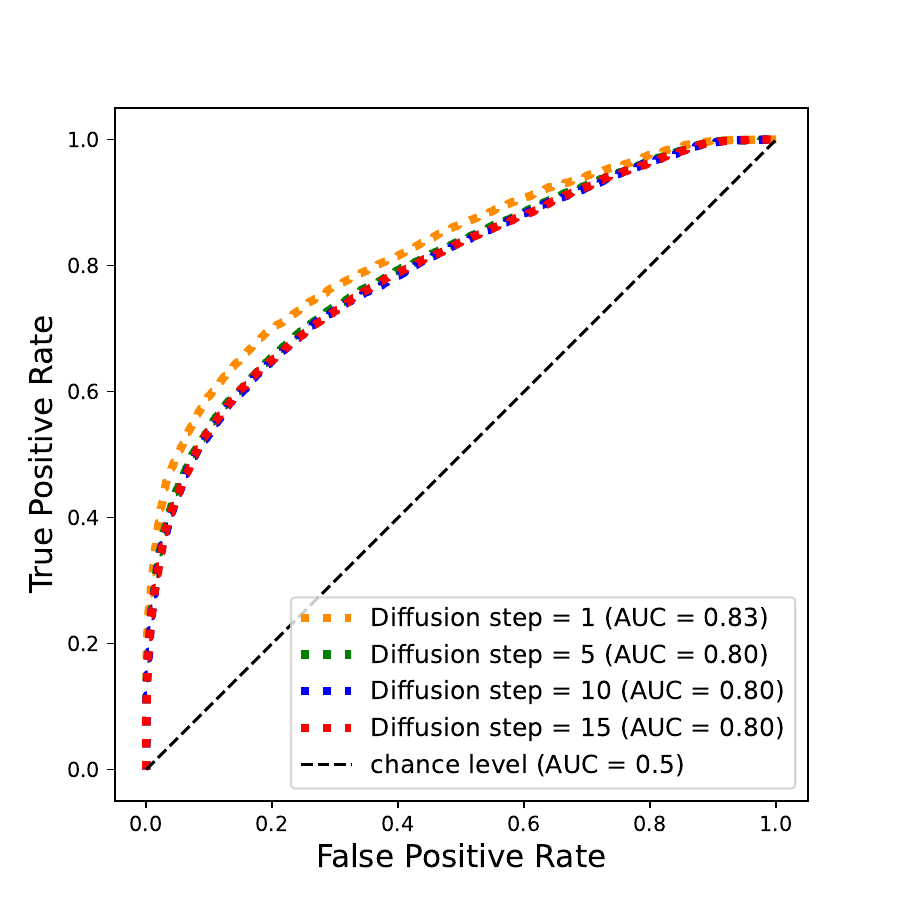}
         \caption{Elastic net}
         \label{fig:provax}
     \end{subfigure}
     \begin{subfigure}[b]{0.43\textwidth}
         \centering
         \includegraphics[width=\textwidth]{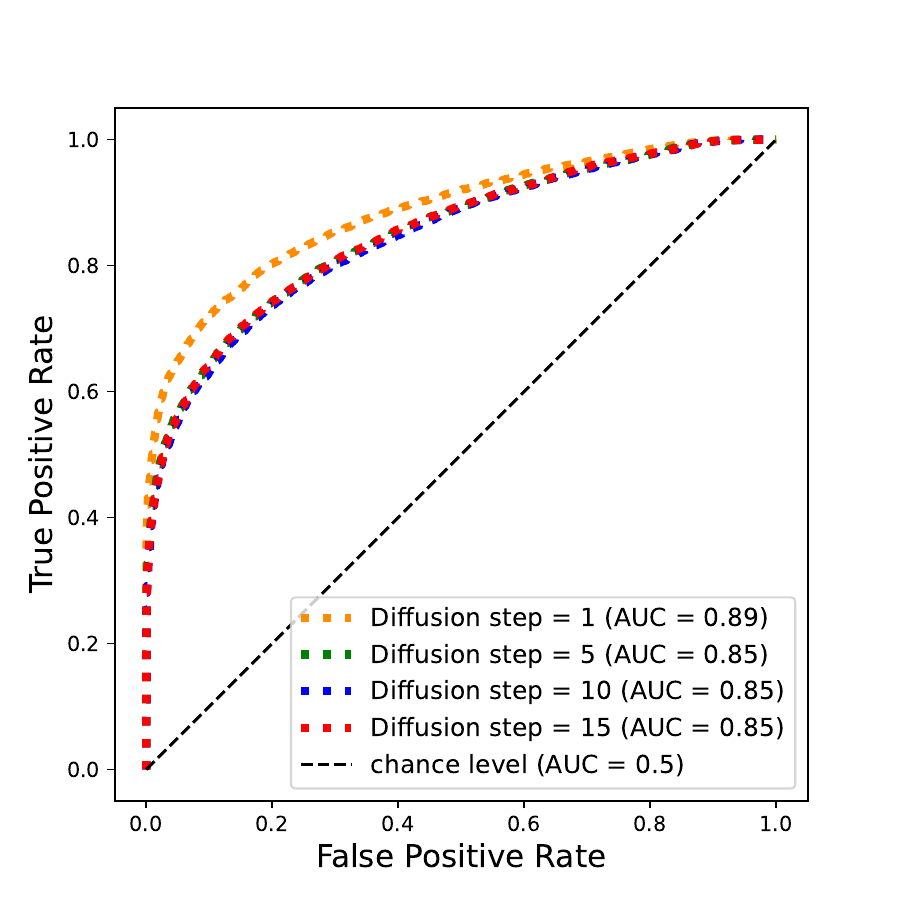}
         \caption{Deep fool}
         \label{fig:prova}
     \end{subfigure}
        \caption{Attack detection ROC curves for DiffDefense. In our experiments FGSM, PGD, EOT+PGD, AutoAttack, Square Attack yielded a AUC $\in$ [.99, 1].}
        \label{fig:roccurve}
\vspace{-15pt}

\end{figure}

Interestingly the results of our study indicate that non-perturbed images are reconstructed with greater ease in comparison to those subjected to adversarial attacks. This is expected since the diffusion model and the classifier are trained on the same data, facilitating the diffusion in the reverse process phase using an unperturbed image to an adversarial image. This ease of reconstruction is reflected in significantly smaller reconstruction errors after an equal number of iterations. These findings suggest that the reconstruction error may serve as a potential indicator of the presence of an attack. In Fig.~\ref{fig:roccurve} we show ROC curves varying the diffusion step for \cite{elasticnet, deepfool}. For all other methods\cite{pgd,eotpgd, autoattack, squarea} we get AUC $\in$ [.99, 1].

\vspace{-25pt}
\section{Conclusion}

We proposed DiffDefense, a novel method that uses Diffusion models for reconstruction, enhancing classifier robustness against attacks. Empirical evaluation demonstrated its efficacy, speed, and potential as an alternative to GAN-based methods and adversarial purification methods based on diffusion models. We also showed that our approach is effective against previously unseen attacks, highlighting its robustness to new attacks. Additionally, we illustrated the usefulness of reconstruction as a tool for adversarial detection.
Our findings suggest that Diffusion based adversarial defense by reconstruction is a promising path toward developing secure AI systems. We believe that future work may further improve our method by adopting better solvers for more accurate and faster reconstruction. 

\vspace{-10pt}
\bibliographystyle{splncs04}
\bibliography{diffdef}

\end{document}